\documentclass{article}

\usepackage[main, final]{neurips_2026}

\usepackage[utf8]{inputenc}
\usepackage[T1]{fontenc}
\usepackage[hidelinks]{hyperref}
\usepackage{url}
\usepackage{booktabs}
\usepackage{amsfonts}
\usepackage{amsmath}
\usepackage{nicefrac}
\usepackage{microtype}
\usepackage{xcolor}
\usepackage{graphicx}
\usepackage{multirow}
\usepackage{array}
\title{How Many Labels Does a Language Need? Annotation Budgets and Cross-Lingual Pooling for African-Language Text Classification}

\author{%
  Bhanu Prakash Vangala \\
  University of Missouri \\
  \texttt{bv3hz@missouri.edu} \And
  Sowmya Guda \\
  University of Missouri \\
  \texttt{sghmy@missouri.edu} \AND
  Latha Peddi \\
  Government Degree College \\
  \texttt{lathakappala@outlook.com} \And
  Navya Vangala \\
  Amar Bio Tech Pvt Ltd \\
  \texttt{vangalanavya.8@gmail.com}
}

\begin{document}

\maketitle

\begin{abstract}
Every text classifier for an African language begins with a budgeting question: how many labelled examples are needed, and can labels from other African languages stand in for them? We answer both questions empirically for 28 language-task pairs, news topic classification in 16 languages (MasakhaNEWS) and tweet sentiment in 12 languages (AfriSenti), using a character n-gram linear model that trains in seconds on two CPU cores with no pretrained weights and no accelerator. Monolingual learning curves at budgets from 25 to several thousand labels show that topic classification reaches 90\% of its full-data macro-F1 with about 400 labels in the median language, while sentiment is still improving at the full training size in 11 of 12 languages and needs thousands of labels. Pooling the full training data of the other languages in the benchmark is worth a great deal at small budgets and nothing at large ones: at 25 target labels it adds 0.20 macro-F1 on average for news (up to 0.43 for Lingala) and 0.08 for sentiment, the gain decays to zero by 800 labels, and at full size pooling hurts in 9 of 16 and 8 of 12 languages. Twenty-five target labels plus pooled data match what 100 to 400 monolingual labels achieve for most news languages. A complete zero-shot transfer matrix shows that transfer without any target labels recovers a median of only 13\% (news) and 4\% (sentiment) of the gap between a majority-class predictor and the in-language model, with the exceptions explained by shared script (Amharic and Tigrinya), shared lexicon (English and Nigerian Pidgin, the Arabic dialects), or a shared label prior rather than by language family. We release code that regenerates every number from the public benchmark files and translate the results into concrete annotation guidance for teams building African-language classifiers without GPUs.
\end{abstract}

\section{Introduction}

Every text classifier for an African language starts with the same question: how many labelled examples do we need? The answer decides whether a newsroom, a health hotline, or a moderation team can afford to build the system at all. Annotation is the dominant cost of African-language NLP. It is paid by native speakers whose time is scarce, and it is paid again for every new language and every new task. Yet the question is rarely answered directly. Benchmark papers report performance at the full training size they happened to collect, and few-shot studies report performance at a handful of examples with large pretrained models that most deployments cannot run. Between those two points, where real projects live, there is little guidance.

This paper measures the answer for 28 language-task pairs across two public benchmarks, MasakhaNEWS \citep{adelani2023masakhanews} for news topic classification in 16 languages and AfriSenti \citep{muhammad2023afrisenti, muhammad2023semeval} for tweet sentiment in 12 languages. We train a single, deliberately simple model family, character n-gram TF-IDF with a linear support vector machine, at annotation budgets from 25 to several thousand labelled examples per language, and we record macro-F1 at every budget. The model needs no accelerator and no pretrained weights. It trains in seconds on a laptop, which is the compute that most African deployments actually have \citep{ahia2021double, leong2023adapting}.

We then ask the second question that every project with a small budget asks: can labelled data from other African languages substitute for labels in ours? We measure this two ways. Pooling adds the full training data of all other languages in the benchmark to the target language's small budget and trains one shared model. Zero-shot transfer trains on one language and tests on another, for every ordered pair, giving a transfer matrix across scripts and language families. Both are cheap to run and both are the kind of thing a team would try before spending money on annotation.

The findings are practical. For news topic classification, a few hundred labels in the target language are enough for a simple model to reach most of its full-data performance, and pooling the other languages' data lets a team start with a few dozen. For sentiment, no budget in the benchmark range is enough, pooling helps only the smallest budgets, and at full size it hurts most languages. Zero-shot transfer between African languages, the cheapest option of all, mostly does not work, and where it appears to work the reason is shared script, shared vocabulary, or a shared label distribution rather than linguistic relatedness.

Our contributions are a set of measured learning curves for 28 African-language classification tasks under a compute budget any team can meet, a per-language estimate of the label budget at which a simple model reaches most of its full-data performance, a controlled measurement of when pooling other African languages' data helps and when it hurts, a complete zero-shot transfer matrix that separates script sharing from genuine linguistic transfer, and a release of code that regenerates every number from the public benchmark files.

\section{Related Work}

\paragraph{African-language benchmarks and models.}
Community efforts, most visibly Masakhane \citep{nekoto2020participatory}, have produced benchmark datasets for named entity recognition \citep{adelani2021masakhaner}, news topic classification \citep{adelani2023masakhanews}, and sentiment \citep{muhammad2023afrisenti}, and pretrained encoders adapted to African languages such as AfriBERTa \citep{ogueji2021small} and AfroXLMR \citep{alabi2022adapting}, built on XLM-R \citep{conneau2020unsupervised}. These papers report strong results at full training size. The MasakhaNEWS paper also reports few-shot results with pattern-exploiting training \citep{schick2021exploiting}, reaching about 93\% of full-supervision performance with ten examples per label, which is an important data point but one that requires a large pretrained model and GPU fine-tuning. \citet{joshi2020state} and \citet{adebara2022towards} document how little of NLP reaches African languages, and \citet{kreutzer2022quality} show that the web-crawled data behind multilingual pretraining is often unusable for exactly these languages.

\paragraph{Low compute.}
\citet{ahia2021double} name the low-resource double bind: the languages with the least data are also served by the communities with the least compute, so methods that assume large models compound the disadvantage. \citet{leong2023adapting} evaluate low-compute adaptation methods on African languages and find that simple approaches remain competitive. \citet{hooker2021hardware} and \citet{schwartz2020green} make the general case for reporting cost. Character n-gram models \citep{cavnar1994ngram} and linear classifiers over sparse features \citep{joulin2017bag} are the standard low-compute baseline; we use them because they are what a team without a GPU will use, and because their learning curves are easy to interpret.

\paragraph{Learning curves and transfer.}
Sample-size planning through learning curves has a long history in applied classification \citep{figueroa2012predicting}, and active learning \citep{settles2009active} exists to reduce the budget further. In multilingual NLP, \citet{lauscher2020zero} show that zero-shot transfer from multilingual transformers degrades sharply for distant, low-resource target languages, and that a few target labels recover most of the loss. \citet{hedderich2020transfer} study transfer and distant supervision for Hausa and Yorùbá and find that small amounts of target data matter more than the transfer source. \citet{lin2019choosing} learn to rank transfer languages from linguistic and dataset features. \citet{hu2020xtreme} standardise cross-lingual evaluation for high-resource models. Our study is the low-compute complement to this line: no pretrained model, no GPU, and an explicit account of the label budget in each language.

\section{Setup}
\label{sec:setup}

\paragraph{Data.}
MasakhaNEWS \citep{adelani2023masakhanews} covers Amharic, English, French, Hausa, Igbo, Lingala, Luganda, Oromo, Nigerian Pidgin, Rundi, chiShona, Somali, Swahili, Tigrinya, isiXhosa, and Yorùbá, with four to seven topic labels per language. Label sets differ across languages: Amharic has four topics, Hausa seven. Training sets range from 608 (Lingala) to 3{,}309 (English) articles. We classify the headline concatenated with the first 1{,}000 characters of the body. AfriSenti \citep{muhammad2023afrisenti} covers the 12 languages of the SemEval-2023 Task 12 monolingual track (Algerian Arabic, Amharic, Hausa, Igbo, Kinyarwanda, Moroccan Arabic, Mozambican Portuguese, Nigerian Pidgin, Swahili, Xitsonga, Twi, Yorùbá) with three sentiment labels; training sets range from 804 (Xitsonga) to 14{,}172 (Hausa) tweets. We use the official train and test splits as distributed in the benchmark repositories and report macro-F1, the metric both benchmarks use, on the official test split.

\paragraph{Model.}
Character n-grams of length 2 to 5 within word boundaries, TF-IDF weighted with sublinear term frequency and capped at 200{,}000 features, followed by a linear support vector machine ($C = 0.5$) from scikit-learn \citep{pedregosa2011scikit}. Character n-grams handle the Ge'ez, Arabic, and Latin scripts in the data without any tokeniser, normaliser, or language-specific resource, which is what makes the same pipeline usable for every language. The vectoriser is fit on whatever training data the configuration uses, so a 25-example run sees only those 25 examples.

\paragraph{Configurations.}
For each language we draw stratified subsamples of the official training set at budgets $n \in \{25, 50, 100, 200, 400, 800\}$ and also use the full training set. Each budget below the full size is drawn with three random seeds. Two settings are trained at every budget. \emph{Monolingual} trains on the $n$ target-language examples alone. \emph{Pooled} trains on the same $n$ examples together with the full training sets of every other language in the benchmark, with no language tag and no reweighting; the vectoriser is fit on the other languages' data and applied to the target examples. We also train the pooled model with zero target examples, which is the multilingual zero-shot setting. For the transfer matrix we train on the full training set of each source language and test on the official test set of every target language in the same benchmark. Predictions of labels that do not exist in a target language count as errors. Every run records wall-clock training time and the size of the weight matrix, all on two CPU cores with no accelerator.

\paragraph{Label budget estimate.}
Following \citet{figueroa2012predicting} we fit an inverse power law $F(n) = a - b\,n^{-c}$ to each monolingual learning curve and report $n_{95}$, the smallest budget at which the fitted curve reaches 95\% of the full-data macro-F1. We also report the raw budget at which the measured mean first reaches that level, which needs no model.

\section{Results}
\label{sec:results}

\subsection{Monolingual learning curves and label budgets}
\label{sec:curves}

\begin{figure}[tb]
  \centering
  \includegraphics[width=\linewidth]{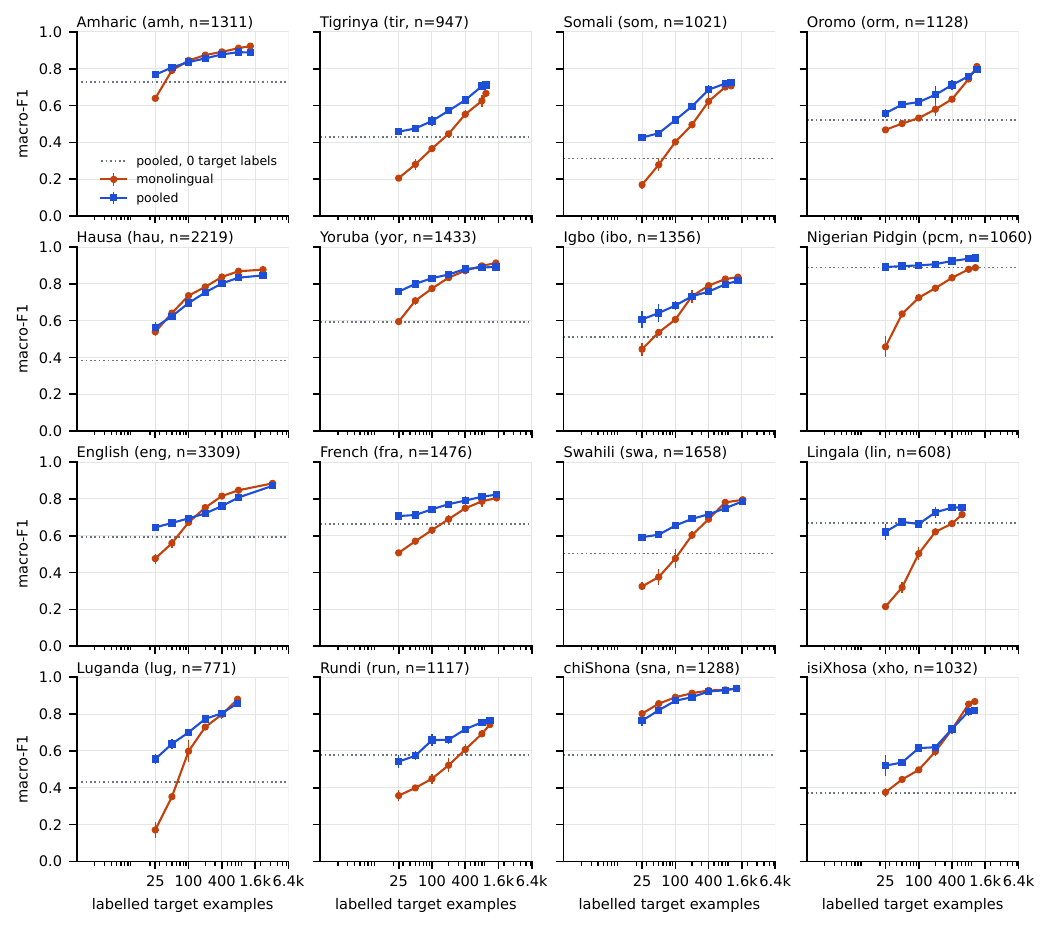}
  \caption{MasakhaNEWS learning curves. Macro-F1 on the official test split against the number of labelled target-language training examples (log scale), for a model trained on those examples alone (monolingual) and for one trained on those examples plus the full training sets of the other 15 languages (pooled). Bars are $\pm 1$ standard deviation over three subsamples. The dotted line is the pooled model with zero target examples.}
  \label{fig:curves-mn}
\end{figure}

\begin{figure}[tb]
  \centering
  \includegraphics[width=\linewidth]{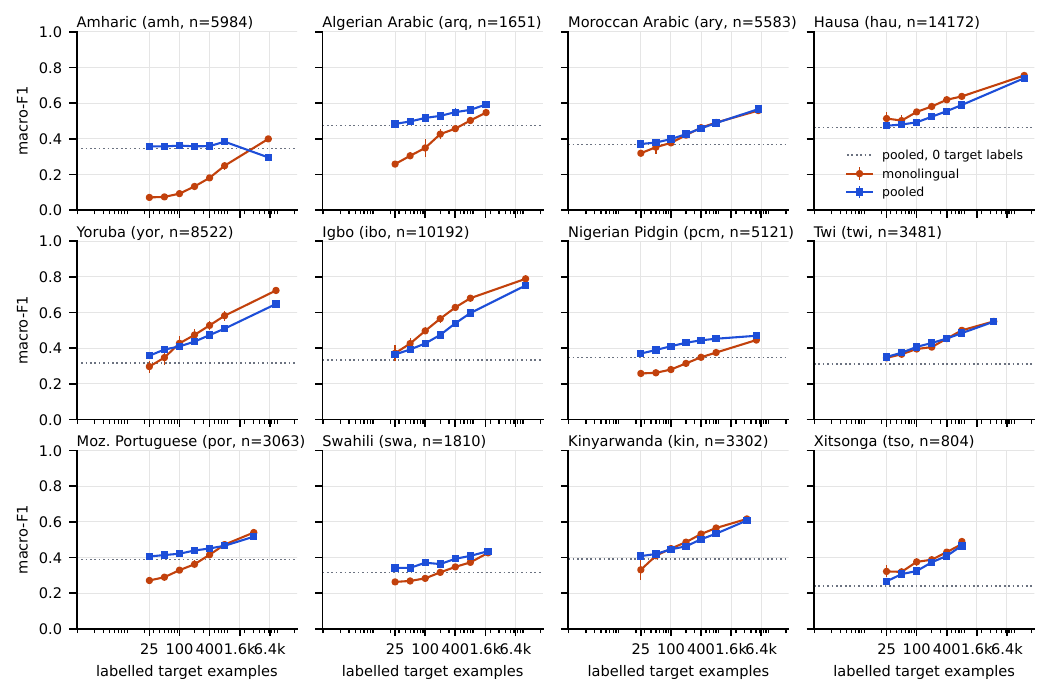}
  \caption{AfriSenti learning curves, same layout as Figure~\ref{fig:curves-mn}. Sentiment is far from saturated at the full training size in every language except Amharic, whose curve is dominated by a train-test label shift (52\% neutral in training, 67\% negative in test).}
  \label{fig:curves-as}
\end{figure}

Figures~\ref{fig:curves-mn} and~\ref{fig:curves-as} show the monolingual curves, and Table~\ref{tab:summary} gives the numbers. For news topic classification the curves rise steeply and then flatten. With 100 labels the monolingual model reaches on average 72\% of its full-data macro-F1; with 400 labels it reaches 90\%. The raw budget at which a language first reaches 95\% of its full-data score, $n_{95}$, ranges from 200 (chiShona) to the full training set (Tigrinya, Oromo, Rundi, Lingala, and Luganda, whose curves are still rising), with a median of 800. The fitted power law gives a median $n_{95}$ of 575 and puts most languages between 400 and 900. These are budgets a small team can meet: a native-speaking annotator labelling news headlines and leads at a few per minute produces 500 labels in a working day.

Sentiment is a different problem. With 100 labels the monolingual model reaches 63\% of its full-data score and with 400 labels 78\%, but the full-data score itself is low (0.40 to 0.79 macro-F1) and the curves have not flattened. In 11 of 12 languages the raw $n_{95}$ is the full training set, and the fitted estimates range from about 700 (Xitsonga) to 8{,}000 (Amharic) with a median near 2{,}300. Sentiment on short, noisy, code-switched social-media text needs an order of magnitude more labels than news topic classification, and a character n-gram model may never close the gap; this is where pretrained representations earn their compute. The Amharic curve also shows a hazard that budget planning must account for: the training and test label distributions differ sharply (52\% neutral in training against 67\% negative in test), so a small model that learns the training prior scores near zero on the test set until it has seen enough data to learn actual sentiment cues.

\begin{table}[tb]
  \caption{Macro-F1 on the official test split. $n$ is the full training size. \emph{zero} is the pooled model with no target labels; \emph{m} and \emph{p} are the monolingual and pooled models at 100 and 400 target labels; \emph{full} is the model trained on all target labels, alone or pooled. $n_{95}$ is the smallest budget at which the monolingual model reaches 95\% of its full-data score, measured (raw) and from a fitted power law (fit); a raw value equal to $n$ means the curve had not flattened.}
  \label{tab:summary}
  \centering
  \scriptsize
  \setlength{\tabcolsep}{4pt}
  \begin{tabular}{lr c c cc cc c cc}
    \toprule
    & & \multicolumn{2}{c}{full mono} & \multicolumn{2}{c}{100 labels} & \multicolumn{2}{c}{400 labels} & full & \multicolumn{2}{c}{$n_{95}$} \\
    \cmidrule(lr){3-4} \cmidrule(lr){5-6} \cmidrule(lr){7-8} \cmidrule(lr){10-11}
    language & $n$ & mono & zero & m & p & m & p & pooled & raw & fit \\
    \midrule
    \multicolumn{11}{l}{\emph{MasakhaNEWS}} \\
    Amharic & 1311 & 0.923 & 0.727 & 0.845 & 0.836 & 0.892 & 0.878 & 0.888 & 400 & 181 \\
    Tigrinya & 947 & 0.666 & 0.430 & 0.366 & 0.516 & 0.552 & 0.630 & 0.711 & 947 & 866 \\
    Somali & 1021 & 0.707 & 0.312 & 0.402 & 0.521 & 0.623 & 0.686 & 0.726 & 800 & 685 \\
    Oromo & 1128 & 0.812 & 0.520 & 0.532 & 0.619 & 0.634 & 0.712 & 0.794 & 1128 & 1463 \\
    Hausa & 2219 & 0.877 & 0.384 & 0.736 & 0.696 & 0.837 & 0.802 & 0.846 & 400 & 427 \\
    Yoruba & 1433 & 0.913 & 0.592 & 0.775 & 0.831 & 0.871 & 0.881 & 0.892 & 400 & 373 \\
    Igbo & 1356 & 0.836 & 0.513 & 0.606 & 0.682 & 0.790 & 0.758 & 0.817 & 800 & 542 \\
    Nigerian Pidgin & 1060 & 0.888 & 0.888 & 0.725 & 0.900 & 0.833 & 0.924 & 0.941 & 800 & 423 \\
    English & 3309 & 0.885 & 0.593 & 0.673 & 0.695 & 0.816 & 0.761 & 0.871 & 800 & 788 \\
    French & 1476 & 0.805 & 0.663 & 0.631 & 0.744 & 0.749 & 0.791 & 0.823 & 800 & 608 \\
    Swahili & 1658 & 0.796 & 0.505 & 0.477 & 0.656 & 0.690 & 0.717 & 0.787 & 800 & 848 \\
    Lingala & 608 & 0.715 & 0.669 & 0.503 & 0.664 & 0.666 & 0.753 & 0.754 & 608 & 407 \\
    Luganda & 771 & 0.880 & 0.431 & 0.598 & 0.701 & 0.798 & 0.804 & 0.858 & 771 & 493 \\
    Rundi & 1117 & 0.743 & 0.577 & 0.449 & 0.658 & 0.607 & 0.718 & 0.767 & 1117 & 1080 \\
    chiShona & 1288 & 0.939 & 0.576 & 0.891 & 0.872 & 0.928 & 0.923 & 0.939 & 200 & 101 \\
    isiXhosa & 1032 & 0.868 & 0.372 & 0.497 & 0.615 & 0.717 & 0.719 & 0.819 & 800 & 936 \\
    \midrule
    \multicolumn{11}{l}{\emph{AfriSenti}} \\
    Amharic & 5984 & 0.400 & 0.345 & 0.093 & 0.362 & 0.182 & 0.359 & 0.296 & 5984 & 8173 \\
    Algerian Arabic & 1651 & 0.548 & 0.476 & 0.349 & 0.517 & 0.457 & 0.550 & 0.593 & 1651 & 1037 \\
    Moroccan Arabic & 5583 & 0.559 & 0.370 & 0.378 & 0.399 & 0.463 & 0.459 & 0.567 & 5583 & 2421 \\
    Hausa & 14172 & 0.756 & 0.465 & 0.551 & 0.493 & 0.620 & 0.556 & 0.741 & 14172 & 5725 \\
    Yoruba & 8522 & 0.724 & 0.316 & 0.427 & 0.410 & 0.528 & 0.474 & 0.648 & 8522 & 4416 \\
    Igbo & 10192 & 0.789 & 0.332 & 0.497 & 0.427 & 0.629 & 0.540 & 0.752 & 10192 & 3544 \\
    Nigerian Pidgin & 5121 & 0.446 & 0.347 & 0.280 & 0.411 & 0.349 & 0.444 & 0.470 & 5121 & 3236 \\
    Twi & 3481 & 0.550 & 0.312 & 0.396 & 0.407 & 0.454 & 0.454 & 0.549 & 3481 & 1926 \\
    Moz. Portuguese & 3063 & 0.540 & 0.388 & 0.329 & 0.421 & 0.415 & 0.451 & 0.517 & 3063 & 2140 \\
    Swahili & 1810 & 0.426 & 0.316 & 0.283 & 0.371 & 0.347 & 0.393 & 0.436 & 1810 & 1680 \\
    Kinyarwanda & 3302 & 0.616 & 0.392 & 0.450 & 0.446 & 0.532 & 0.502 & 0.608 & 3302 & 1372 \\
    Xitsonga & 804 & 0.490 & 0.239 & 0.376 & 0.324 & 0.430 & 0.409 & 0.464 & 800 & 701 \\
    \bottomrule
  \end{tabular}
\end{table}

\subsection{Pooling other languages: large gains at small budgets, none at large ones}
\label{sec:pooling}

\begin{figure}[tb]
  \centering
  \includegraphics[width=\linewidth]{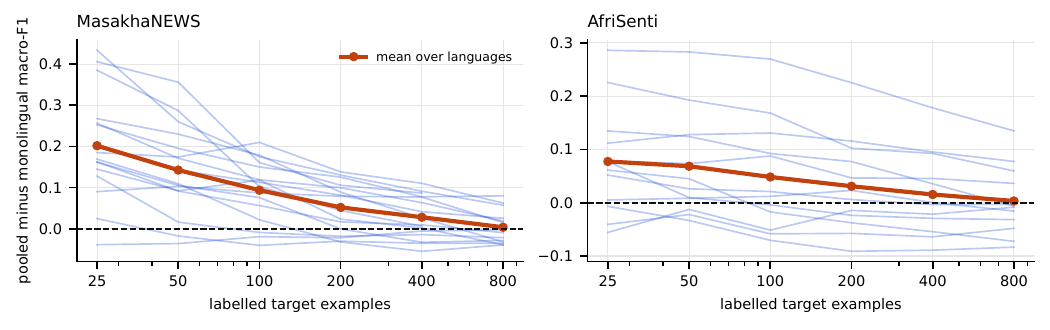}
  \caption{Gain from pooling, pooled minus monolingual macro-F1, as a function of the target-language budget. Thin lines are languages; the thick line is the mean. Pooling is worth about 0.20 macro-F1 at 25 news labels and 0.08 at 25 sentiment labels, and nothing by 800.}
  \label{fig:gain}
\end{figure}

Figure~\ref{fig:gain} shows the difference between the pooled and monolingual curves. At 25 target labels, pooling the other languages' training data raises news macro-F1 by 0.20 on average and by more than 0.35 for Lingala, Luganda, and Nigerian Pidgin; it helps 15 of 16 languages. The gain halves at every doubling of the budget: 0.14 at 50 labels, 0.09 at 100, 0.05 at 200, 0.03 at 400, and 0.004 at 800, where pooling helps only 6 of 16 languages. At the full training size pooling is a wash on average and hurts 9 of 16 languages, by up to 0.05 (isiXhosa) and 0.03 to 0.04 (Amharic, Hausa). The same shape appears for sentiment with smaller amplitude: 0.08 at 25 labels, helping 9 of 12 languages, decaying to zero by 800, and at full size hurting 8 of 12 languages, by as much as 0.10 for Amharic.

The practical reading is a label-equivalent. Twenty-five target labels plus pooled data match what the monolingual model achieves with 100 to 400 labels in 12 of 16 news languages, and with 50 labels for Amharic and Hausa (Table~\ref{tab:summary}); for Nigerian Pidgin, pooled data with 25 labels already matches the full monolingual model, because English is in the pool. For sentiment the equivalent is smaller and less consistent, from 25 labels (no gain) for Hausa, Igbo, and Xitsonga to 800 for Algerian Arabic and Nigerian Pidgin, and the full set for Amharic. Pooling is therefore the right first step for a team with a few dozen labels and the wrong last step for a team with a few thousand, and the crossover is at a few hundred labels for both tasks.

Why pooling stops helping is visible in the transfer matrix below. The shared model has to fit every language's decision boundary with one weight vector over a shared character n-gram space; other languages' data sets the boundary for n-grams the target language shares with them, which is helpful when the target has too few examples to set it itself and harmful once it does, because the other languages' usage of those n-grams is not the target's. The pooled model is also five times larger (11 MB against 2 MB of weights for news) and takes 50 seconds rather than 2 to train, which matters little on a laptop but is the wrong direction for a phone.

\subsection{Zero-shot transfer: script and lexicon, not family}
\label{sec:transfer}

\begin{figure}[tb]
  \centering
  \includegraphics[width=\linewidth]{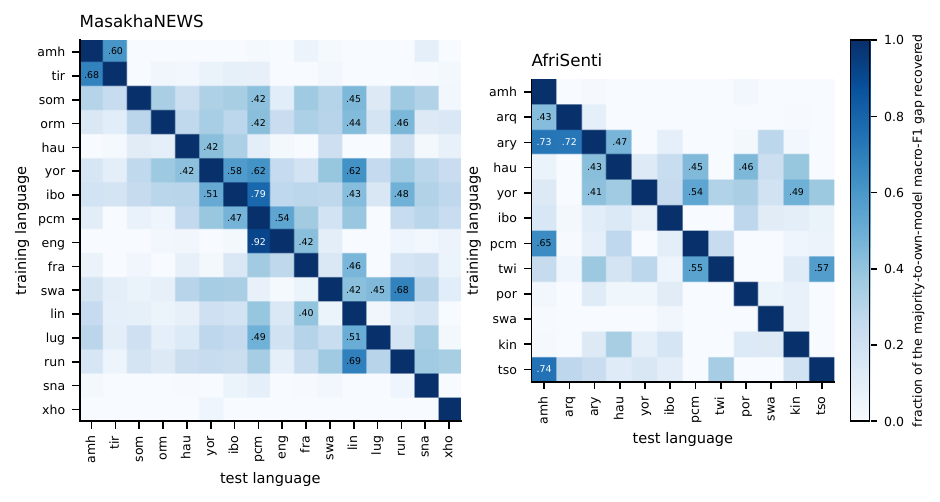}
  \caption{Zero-shot transfer. A model trained on the row language's full training set is tested on the column language. Cells show the fraction of the gap between a majority-class predictor and the column language's own full-data model that the transferred model recovers, so 0 means no better than guessing the most common label and 1 means as good as in-language training. Languages are ordered by script and family: Ge'ez (amh, tir), Cushitic (som, orm), Nigerian languages (hau, yor, ibo, pcm), colonial languages (eng, fra), Bantu (swa, lin, lug, run, sna, xho); Arabic dialects (amh is Ge'ez), Nigerian, and Bantu for AfriSenti.}
  \label{fig:transfer}
\end{figure}

Figure~\ref{fig:transfer} shows the full transfer matrix. The typical off-diagonal cell recovers a median of 13\% of the majority-to-own-model gap for news and 4\% for sentiment; only 12\% and 11\% of the source-target pairs recover more than 40\%. Zero-shot transfer between African languages with a surface-form model is, as a rule, close to useless. The exceptions are informative because each has a mechanism that is not linguistic relatedness. English to Nigerian Pidgin recovers 92\% of the gap and Igbo to Pidgin 79\%, because Pidgin is an English-lexified creole and the Nigerian news sources share named entities. Tigrinya to Amharic recovers 68\% and Amharic to Tigrinya 60\%, the only pair in the data sharing the Ge'ez script, so their character n-grams overlap heavily. Rundi to Lingala (69\%) and Swahili to Rundi (68\%) are Bantu pairs with shared loanwords and place names. For sentiment, the top pairs are Moroccan to Algerian Arabic (72\%), the two Arabic-script dialects, and a cluster of sources that transfer to Amharic (Xitsonga 74\%, Moroccan Arabic 73\%, Nigerian Pidgin 65\%) which is a label-prior artefact: those models predict negative for most Amharic tweets, and 67\% of the Amharic test set is negative. Family membership on its own, for instance Hausa to Somali or Oromo (both Afroasiatic but different branches), buys nothing.

This matters for how teams choose a transfer source. The linguistic-similarity heuristics that work for pretrained multilingual encoders \citep{lin2019choosing} do not carry over to the surface-form models that run on a CPU. For these models the useful predictors are shared script, shared lexicon (colonial and creole relationships, dialect continua), and shared named entities from a shared news ecosystem, and the label prior must be checked before any transfer number is believed.

\subsection{Cost}
\label{sec:cost}

The full monolingual model trains in 0.1 to 6 seconds per language and the pooled model in 18 to 53 seconds, on two CPU cores with 7 GB of memory. Scoring a test set of a few hundred to a few thousand documents takes under a second. Generating every number in this paper, 28 languages at seven budgets in two settings with three seeds plus the two transfer matrices, took about two and a half hours on the same two cores. No GPU, no pretrained model, and no external service were used at any point. We report this not as a limitation but as the operating point of the study: it is the compute that a newsroom, an NGO, or a university lab in most of Africa can rely on \citep{ahia2021double, hooker2021hardware}, and the learning curves above are the curves that compute produces.

\section{Discussion}
\label{sec:discussion}

\paragraph{Annotation guidance.}
The results translate into budgets. For a news-style topic task, plan on 400 to 800 labels per language for a simple model to reach 90\% to 95\% of what it will ever do, and start with pooled data from any other African-language news corpus with the same label scheme, which makes the first 25 to 100 labels worth several times their count. Drop the pooled data once the target budget passes a few hundred labels, since it then costs accuracy as well as compute. For sentiment on social media, do not expect a character n-gram model to be good at any budget; plan on thousands of labels, treat pooling as a small early boost only, and expect to move to a pretrained encoder \citep{alabi2022adapting, ogueji2021small} once labels exist. Before using any transfer source, check its label prior against the target's, because a model that recovers 70\% of the gap on paper may be doing nothing more than predicting the majority class.

\paragraph{What the curves do not show.}
Macro-F1 on the official test split is one number per language, and the official splits carry their own histories: the Amharic AfriSenti shift shows that a test set can differ from its training set in ways that dominate a small model's score. Three subsamples per budget give error bars but not a full picture of variance at 25 labels, where a single unlucky draw can miss a class entirely. We used one model family, chosen because it is what runs without a GPU; a pretrained encoder would shift every curve up and, from the evidence in \citet{adelani2023masakhanews}, would flatten the news curves much earlier. We did not study active learning \citep{settles2009active}, which could reduce the budgets we report, nor distant supervision \citep{hedderich2020transfer}. The pooled setting mixes languages without weighting; language-tagged or upweighted variants might extend the range over which pooling helps.

\paragraph{Relation to prior findings.}
Our curves agree with \citet{lauscher2020zero} and \citet{hedderich2020transfer} that a few target labels are worth more than any transfer source, and extend the finding to a regime those papers do not cover: no pretrained model at all. The MasakhaNEWS few-shot result, about 93\% of full performance from ten examples per label with a large model and pattern-exploiting training, sits at the same 90\% mark that our simple model reaches with 400 labels. The choice between the two is a choice between 40 labels and a GPU or 400 labels and a laptop, and for most African deployments the second is the one that exists.

\section{Conclusion}

Across 28 African-language classification tasks, a character n-gram linear model with no pretrained weights reaches most of its full-data performance with a few hundred labels for news topics and needs thousands for tweet sentiment. Pooling the training data of other African languages is worth up to 0.4 macro-F1 when the target has a few dozen labels and costs accuracy once it has a few hundred, with the crossover at a few hundred labels in both tasks. Zero-shot transfer without target labels recovers little of the in-language performance except where languages share a script, a lexicon, or a label prior. All of it runs on two CPU cores in an afternoon, and all of it can be regenerated from the public benchmark files with the released code. The numbers are the budget a team should plan for, measured under the compute that team has.

{\small
\bibliographystyle{plainnat}
\bibliography{refs}

@inproceedings{adelani2023masakhanews,
  title={{MasakhaNEWS}: News topic classification for {African} languages},
  author={Adelani, David Ifeoluwa and Masiak, Marek and Azime, Israel Abebe and Alabi, Jesujoba and Tonja, Atnafu Lambebo and Mwase, Christine and Ogundepo, Odunayo and Dossou, Bonaventure F. P. and Oladipo, Akintunde and Nixdorf, Doreen and others},
  booktitle={Proceedings of the 13th International Joint Conference on Natural Language Processing and the 3rd Conference of the Asia-Pacific Chapter of the Association for Computational Linguistics (IJCNLP-AACL)},
  year={2023}
}

@inproceedings{muhammad2023afrisenti,
  title={{AfriSenti}: A {Twitter} sentiment analysis benchmark for {African} languages},
  author={Muhammad, Shamsuddeen Hassan and Abdulmumin, Idris and Ayele, Abinew Ali and Ousidhoum, Nedjma and Adelani, David Ifeoluwa and Yimam, Seid Muhie and Ahmad, Ibrahim Sa'id and Beloucif, Meriem and Mohammad, Saif M. and Ruder, Sebastian and others},
  booktitle={Proceedings of the 2023 Conference on Empirical Methods in Natural Language Processing (EMNLP)},
  year={2023}
}

@inproceedings{muhammad2023semeval,
  title={{SemEval}-2023 {Task} 12: Sentiment analysis for {African} languages ({AfriSenti-SemEval})},
  author={Muhammad, Shamsuddeen Hassan and Abdulmumin, Idris and Yimam, Seid Muhie and Adelani, David Ifeoluwa and Ahmad, Ibrahim Sa'id and Ousidhoum, Nedjma and Ayele, Abinew Ali and Mohammad, Saif M. and Beloucif, Meriem and Ruder, Sebastian},
  booktitle={Proceedings of the 17th International Workshop on Semantic Evaluation (SemEval-2023)},
  year={2023}
}

@inproceedings{nekoto2020participatory,
  title={Participatory research for low-resourced machine translation: A case study in {African} languages},
  author={Nekoto, Wilhelmina and Marivate, Vukosi and Matsila, Tshinondiwa and Fasubaa, Timi and Fagbohungbe, Taiwo and Akinola, Solomon Oluwole and Muhammad, Shamsuddeen and Kabongo Kabenamualu, Salomon and Osei, Salomey and Sackey, Freshia and others},
  booktitle={Findings of the Association for Computational Linguistics: EMNLP 2020},
  year={2020}
}

@inproceedings{joshi2020state,
  title={The state and fate of linguistic diversity and inclusion in the {NLP} world},
  author={Joshi, Pratik and Santy, Sebastin and Budhiraja, Amar and Bali, Kalika and Choudhury, Monojit},
  booktitle={Proceedings of the 58th Annual Meeting of the Association for Computational Linguistics (ACL)},
  year={2020}
}

@inproceedings{ahia2021double,
  title={The low-resource double bind: An empirical study of pruning for low-resource machine translation},
  author={Ahia, Orevaoghene and Kreutzer, Julia and Hooker, Sara},
  booktitle={Findings of the Association for Computational Linguistics: EMNLP 2021},
  year={2021}
}

@inproceedings{leong2023adapting,
  title={Adapting to the low-resource double-bind: Investigating low-compute methods on low-resource {African} languages},
  author={Leong, Colin and Shandilya, Herumb and Dossou, Bonaventure F. P. and Tonja, Atnafu Lambebo and Mathew, Joel and Omotayo, Abdul-Hakeem and Yousuf, Oreen and Akinjobi, Zainab and Emezue, Chris Chinenye and Muhammad, Shamsudeen and others},
  booktitle={4th Workshop on African Natural Language Processing (AfricaNLP), ICLR},
  year={2023}
}

@inproceedings{alabi2022adapting,
  title={Adapting pre-trained language models to {African} languages via multilingual adaptive fine-tuning},
  author={Alabi, Jesujoba O. and Adelani, David Ifeoluwa and Mosbach, Marius and Klakow, Dietrich},
  booktitle={Proceedings of the 29th International Conference on Computational Linguistics (COLING)},
  year={2022}
}

@inproceedings{ogueji2021small,
  title={Small data? No problem! {Exploring} the viability of pretrained multilingual language models for low-resourced languages},
  author={Ogueji, Kelechi and Zhu, Yuxin and Lin, Jimmy},
  booktitle={Proceedings of the 1st Workshop on Multilingual Representation Learning (MRL)},
  year={2021}
}

@inproceedings{conneau2020unsupervised,
  title={Unsupervised cross-lingual representation learning at scale},
  author={Conneau, Alexis and Khandelwal, Kartikay and Goyal, Naman and Chaudhary, Vishrav and Wenzek, Guillaume and Guzm{\'a}n, Francisco and Grave, Edouard and Ott, Myle and Zettlemoyer, Luke and Stoyanov, Veselin},
  booktitle={Proceedings of the 58th Annual Meeting of the Association for Computational Linguistics (ACL)},
  year={2020}
}

@article{hooker2021hardware,
  title={The hardware lottery},
  author={Hooker, Sara},
  journal={Communications of the ACM},
  volume={64},
  number={12},
  pages={58--65},
  year={2021}
}

@article{schwartz2020green,
  title={Green {AI}},
  author={Schwartz, Roy and Dodge, Jesse and Smith, Noah A. and Etzioni, Oren},
  journal={Communications of the ACM},
  volume={63},
  number={12},
  pages={54--63},
  year={2020}
}

@article{pedregosa2011scikit,
  title={Scikit-learn: Machine learning in {Python}},
  author={Pedregosa, Fabian and Varoquaux, Ga{\"e}l and Gramfort, Alexandre and Michel, Vincent and Thirion, Bertrand and Grisel, Olivier and Blondel, Mathieu and Prettenhofer, Peter and Weiss, Ron and Dubourg, Vincent and others},
  journal={Journal of Machine Learning Research},
  volume={12},
  pages={2825--2830},
  year={2011}
}

@inproceedings{lauscher2020zero,
  title={From zero to hero: On the limitations of zero-shot language transfer with multilingual transformers},
  author={Lauscher, Anne and Ravishankar, Vinit and Vuli{\'c}, Ivan and Glava{\v{s}}, Goran},
  booktitle={Proceedings of the 2020 Conference on Empirical Methods in Natural Language Processing (EMNLP)},
  year={2020}
}

@inproceedings{hedderich2020transfer,
  title={Transfer learning and distant supervision for multilingual transformer models: A study on {African} languages},
  author={Hedderich, Michael A. and Adelani, David and Zhu, Dawei and Alabi, Jesujoba and Markus, Udia and Klakow, Dietrich},
  booktitle={Proceedings of the 2020 Conference on Empirical Methods in Natural Language Processing (EMNLP)},
  year={2020}
}

@inproceedings{lin2019choosing,
  title={Choosing transfer languages for cross-lingual learning},
  author={Lin, Yu-Hsiang and Chen, Chian-Yu and Lee, Jean and Li, Zirui and Zhang, Yuyan and Xia, Mengzhou and Rijhwani, Shruti and He, Junxian and Zhang, Zhisong and Ma, Xuezhe and Anastasopoulos, Antonios and Littell, Patrick and Neubig, Graham},
  booktitle={Proceedings of the 57th Annual Meeting of the Association for Computational Linguistics (ACL)},
  year={2019}
}

@article{figueroa2012predicting,
  title={Predicting sample size required for classification performance},
  author={Figueroa, Rosa L. and Zeng-Treitler, Qing and Kandula, Sasikiran and Ngo, Long H.},
  journal={BMC Medical Informatics and Decision Making},
  volume={12},
  number={8},
  year={2012}
}

@inproceedings{joulin2017bag,
  title={Bag of tricks for efficient text classification},
  author={Joulin, Armand and Grave, Edouard and Bojanowski, Piotr and Mikolov, Tomas},
  booktitle={Proceedings of the 15th Conference of the European Chapter of the Association for Computational Linguistics (EACL)},
  year={2017}
}

@inproceedings{cavnar1994ngram,
  title={N-gram-based text categorization},
  author={Cavnar, William B. and Trenkle, John M.},
  booktitle={Proceedings of the 3rd Annual Symposium on Document Analysis and Information Retrieval (SDAIR)},
  year={1994}
}

@article{adelani2021masakhaner,
  title={{MasakhaNER}: Named entity recognition for {African} languages},
  author={Adelani, David Ifeoluwa and Abbott, Jade and Neubig, Graham and D'souza, Daniel and Kreutzer, Julia and Lignos, Constantine and Palen-Michel, Chester and Buzaaba, Happy and Rijhwani, Shruti and Ruder, Sebastian and others},
  journal={Transactions of the Association for Computational Linguistics},
  volume={9},
  pages={1116--1131},
  year={2021}
}

@inproceedings{adebara2022towards,
  title={Towards afrocentric {NLP} for {African} languages: Where we are and where we can go},
  author={Adebara, Ife and Abdul-Mageed, Muhammad},
  booktitle={Proceedings of the 60th Annual Meeting of the Association for Computational Linguistics (ACL)},
  year={2022}
}

@article{kreutzer2022quality,
  title={Quality at a glance: An audit of web-crawled multilingual datasets},
  author={Kreutzer, Julia and Caswell, Isaac and Wang, Lisa and Wahab, Ahsan and van Esch, Daan and Ulzii-Orshikh, Nasanbayar and Tapo, Allahsera and Subramani, Nishant and Sokolov, Artem and Sikasote, Claytone and others},
  journal={Transactions of the Association for Computational Linguistics},
  volume={10},
  pages={50--72},
  year={2022}
}

@inproceedings{hu2020xtreme,
  title={{XTREME}: A massively multilingual multi-task benchmark for evaluating cross-lingual generalisation},
  author={Hu, Junjie and Ruder, Sebastian and Siddhant, Aditya and Neubig, Graham and Firat, Orhan and Johnson, Melvin},
  booktitle={International Conference on Machine Learning (ICML)},
  year={2020}
}

@article{settles2009active,
  title={Active learning literature survey},
  author={Settles, Burr},
  journal={Computer Sciences Technical Report 1648, University of Wisconsin--Madison},
  year={2009}
}

@inproceedings{schick2021exploiting,
  title={Exploiting cloze-questions for few-shot text classification and natural language inference},
  author={Schick, Timo and Sch{\"u}tze, Hinrich},
  booktitle={Proceedings of the 16th Conference of the European Chapter of the Association for Computational Linguistics (EACL)},
  year={2021}
}
}

\end{document}